%% file: main.tex
\PassOptionsToPackage{x11names,dvipsnames,table}{xcolor}
\documentclass[10pt,twocolumn,letterpaper]{article}

\usepackage{cvpr}
\usepackage{times}
\usepackage{epsfig}
\usepackage{graphicx}
\usepackage{amsmath}
\usepackage{amssymb}
\usepackage{multirow}
\usepackage[numbers,sort]{natbib}
\usepackage{etoolbox}
\usepackage{enumitem}
\usepackage{color}

\makeatletter
\renewcommand{\paragraph}{%
  \@startsection{paragraph}{4}%
  {\z@}{2.25ex \@plus 1ex \@minus .2ex}{-1em}%
  {\normalfont\normalsize\bfseries}%
}
\makeatother

\definecolor{self_def_color1}{RGB}{231,75,61}

\usepackage{tikz}
\usetikzlibrary{shapes}
\newrobustcmd*{\mytriangle}[1]{\tikz{\filldraw[draw=#1,fill=#1] (0,0) --
(0.2cm,0) -- (0.1cm,0.2cm);}}
\newrobustcmd*{\mysquare}[1]{\tikz{\filldraw[draw=#1,fill=#1] (0,0.16cm) -- (0,0) --
(0.16cm,0) -- (0.16cm,0.16cm);}}
\newrobustcmd*{\mycircle}[1]{\tikz{\fill[draw=#1,fill=#1] (0.5ex, 0.5ex) circle (0.7ex);}}

\usepackage[pagebackref=true,breaklinks=true,letterpaper=true,colorlinks,bookmarks=false]{hyperref}
\definecolor{rowblue}{RGB}{220,230,240}
\definecolor{myorchid}{RGB}{150,10,30}
\definecolor{myb    lue}{RGB}{10,30,250}
\definecolor{mygreen}{RGB}{10,120,10}
\definecolor{fco}{RGB}{236,112,20}

\cvprfinalcopy 

\ifcvprfinal\fi
\begin{document}

\title{Temporally Distributed Networks for Fast Video Semantic Segmentation}

\author{
Ping Hu$^1$,~Fabian Caba Heilbron$^2$,~Oliver Wang$^2$,~Zhe Lin$^2$,~Stan Sclaroff$^1$,~Federico Perazzi$^2$
\and $^1$Boston University~~~~$^2$Adobe Research}

\maketitle

\input{abstract.tex}

\input{introduction.tex}
\input{relatedworks.tex}
\input{method.tex}
\input{experiments.tex}

\input{conclusions.tex}

{\small
\bibliographystyle{ieee_fullname}
\bibliography{egbib}
}

\end{document}

%% file: abstract.tex
\begin{abstract}
We present TDNet, a temporally distributed network designed for fast and accurate video semantic segmentation.
We observe that features extracted from a certain high-level layer of a deep CNN can be approximated by composing features extracted from several shallower sub-networks.
Leveraging the inherent temporal continuity in videos, we distribute these sub-networks over sequential frames.
Therefore, at each time step, we only need to perform a lightweight computation to extract a sub-features group from a single sub-network.
The full features used for segmentation are then recomposed by the application of a novel attention propagation module that compensates for geometry deformation between frames.
A grouped knowledge distillation loss is also introduced to further improve the representation power at both full and sub-feature levels.
Experiments on Cityscapes, CamVid, and NYUD-v2 demonstrate that our method achieves state-of-the-art accuracy with significantly faster speed and lower latency.
\end{abstract}

%% file: introduction.tex
\section{Introduction}
Video semantic segmentation aims to assign pixel-wise semantic labels to video frames.
As an important task for visual understanding, it has attracted more and more attention from the research community~\cite{jin2017video,li2018low,nilsson2018semantic,shelhamer2016clockwork}. 
The recent successes in dense labeling tasks~\cite{chen2018encoder,fu2018dual,li2019expectation,liu2019auto, wu2019wider,zhao2017pyramid,zhang2018context,Zhu_2019_CVPR} have revealed that strong feature representations are critical for accurate segmentation results. 
However, computing strong features typically require deep networks with high computation cost, thus making it challenging for real-world applications like self-driving cars, robot sensing, and augmented-reality, which require both high accuracy \emph{and} low latency. 

\begin{figure}
\centering
\includegraphics[width=1.02\linewidth]{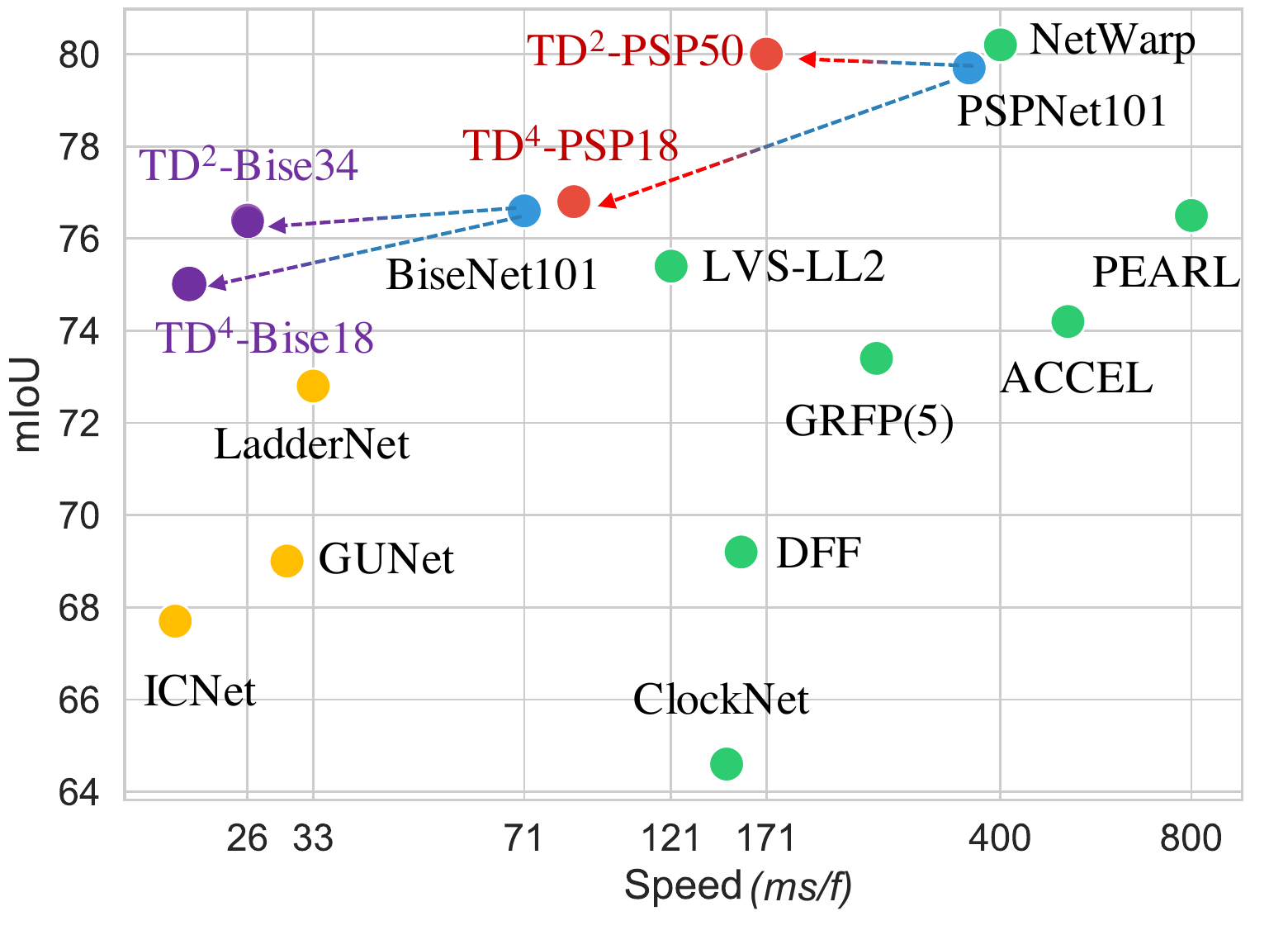}
\vspace{-0.4cm}
    \caption{\small{ Performance on Cityscapes. Our proposed TDNet variants (denoted as \mycircle{self_def_color1} and  \mycircle{Fuchsia}) linked to their corresponding deep image segmentation backbones (denoted as \mycircle{Cerulean}) with similar number of parameters.  
Compared with video semantic segmentation methods NetWarp~\cite{gadde2017semantic}, PEARL~\cite{jin2017video}, ACCEL~\cite{jain2019accel}, LVS-LLS~\cite{li2018low}, GRFP~\cite{nilsson2018semantic}, ClockNet~\cite{shelhamer2016clockwork}, DFF~\cite{zhu2017deep}, and real-time segmentation models LadderNet~\cite{kreso2017ladder}, GUNet~\cite{mazzini2018guided}, and ICNet~\cite{zhao2018icnet}, our TDNet achieves a better balance of accuracy and speed. 
}}
\label{fig1}
\vspace{-0.5cm}
\end{figure}

The most straightforward strategy for video semantic segmentation is to apply a deep image segmentation model to each frame independently, but this strategy does not leverage temporal information provided in the video dynamic scenes. 
One solution, is to apply the same model to all frames and add additional layers on top to model temporal context to extract better features~\cite{gadde2017semantic,jin2017video,kundu2016feature,nilsson2018semantic}.
However, such methods do not help improve efficiency as all features must be recomputed at each frame. 
To reduce redundant computation, a reasonable approach is to apply a strong image segmentation model only at keyframes, and reuse the high-level feature for other frames~\cite{jain2019accel,li2018low,mahasseni2017budget,zhu2017deep}. 
However, the spatial misalignment of other frames with respect to the keyframes is challenging to compensate for and often leads to decreased accuracy comparing to the baseline image segmentation models as reported in~\cite{jain2019accel,li2018low,mahasseni2017budget,zhu2017deep}. 
Additionally, these methods have different computational loads between keyframes and non-keyframes, which results in high maximum latency and unbalanced occupation of computation resources that may decrease system efficiency.

To address these challenges, we propose a novel deep learning model for high-accuracy and low-latency semantic video segmentation named Temporally Distributed Network (TDNet).
Our model is inspired by \textit{Group Convolution}~\cite{ioannou2017deep,krizhevsky2012imagenet}, which shows that extracting features with separated filter groups not only allows for model parallelization, but also helps learn \emph{better} representations.  
Given a deep image segmentation network like PSPNet~\cite{zhao2017pyramid}, we divide the features extracted by the deep model into $N$ (\eg $N$=2 or 4) groups, and use N distinct shallow sub-networks to approximate each group of feature channels. 
By forcing each sub-network to cover a separate feature subspace, a strong feature representation can be produced by reassembling the output of these sub-networks. 
For balanced and efficient computation over time, we let the $N$ sub-networks share the same shallow architecture, which is set to be $\frac{1}{N}$ of the original deep model's size to preserve a similar total model capacity~\cite{wu2019wider,szegedy2017inception,Zagoruyko2016WRN}. 

When segmenting video streams, the $N$ sub-networks are sequentially and circularly assigned to frames over time, such that complementary sub-feature groups are alternatively extracted over time and only one new sub-feature group needs to be computed at each time step. 
To compensate for spatial misalignment caused by motion across frames, we propose an attention propagation module for reassembling features from different time steps. 
To further enhance the network's representational power, we also present a grouped distillation loss to transfer knowledge from a full deep model to our distributed feature network at both full and sub-feature group levels. With this new model, we only need to run a light-weight forward propagation at each frame, and can aggregate full features by \emph{reusing} sub-features extracted in previous frames.
As shown in Fig~\ref{fig1}, our method outperforms state-of-the-art methods while maintaining lower latency.
We validate our approach through extensive experiments over multiple benchmarks. 

In summary, our contributions include: i) a temporally distributed network architecture and grouped knowledge distillation loss that accelerates state-of-the-art semantic segmentation models for videos with more than 2$\times$ lower latency at comparable accuracy; ii) an attention propagation module to efficiently aggregate distributed feature groups over time with robustness to geometry variation across frames; iii) better accuracy and latency than previous state-of-the-art video semantic segmentation methods on three challenging datasets including Cityscapes, Camvid, and NYUD-v2. 

%% file: relatedworks.tex
\section{Related Work}
 Image semantic segmentation is an active area of research that has witnessed significant improvements in performance with the success of deep learning~\cite{liu2019auto,huang2017densely,he2016deep,simonyan2014very}.  
 As a pioneer work, the Fully Convolutional Network (FCN)~\cite{long2015fully} replaced the last fully connected layer for classification with convolutional layers, thus allowing for dense label prediction.
 Based on this formulation, follow-up methods have been proposed for efficient segmentation~\cite{paszke2016enet,zhao2018icnet,yu2018bisenet,li2019dfanet,orsic2019defense,Paul_2020_WACV} or high-quality segmentation~\cite{chen2018encoder,li2019attention,shuai2016dag,TangDPBS18,TangPDASB18,ding2018context,peng2017large,He_2019_ICCV,Takikawa_2019_ICCV,Nekrasov_2020_WACV}. 
 
Semantic segmentation has also been widely applied to videos~\cite{he2017std2p,mahasseni2017budget,tripathi2015semantic,kundu2016feature}, with different approaches employed to balance the trade-off between quality and speed. 
A number of methods leverage temporal context in a video by repeatedly applying the same deep model to each frame and temporally aggregating features with additional network layers~\cite{nilsson2018semantic,gadde2017semantic,jin2017video}. 
Although these methods improve accuracy over single frame approaches, they incur additional computation over a per-frame model.
 
Another group of methods target efficient video segmentation by utilizing temporal continuity to propagate and reuse the high-level features extracted at key frames~\cite{shelhamer2016clockwork,zhu2017deep,jain2019accel,li2018low}. 
The challenge of these methods is how to robustly propagate pixel-level information over time, which might be misaligned due to motion between frames. 
To address this, Shelhamer \etal~\cite{shelhamer2016clockwork} and Carreira  \etal~\cite{carreira2018massively} directly reuse high-level features extracted from deep layers at a low resolution, which they show are relatively stable over time.
Another approach, employed by Zhu \etal~\cite{zhu2017deep} is to adopt optical flow to warp high-level features at keyframes to non keyframes.
Jain~\etal~\cite{jain2019accel} further updates the flow warped feature maps with shallow features extracted at the current frame.
However, using optical flow incurs significant computation cost and can fail with large motion, disocclusions, and non-textured regions. 
To avoid using optical flow, Li \etal~\cite{li2018low} instead proposes to use \textit{spatially variant convolution} to adaptively aggregate features within a local window, which however is still limited by motion beyond that of the predefined window. 
As indicated in~\cite{jain2019accel,li2018low,zhu2017deep}, though the overall computation is reduced compared to their image segmentation baselines, the accuracy is also decreased. In addition, due to the extraction of high-level features at keyframes, these methods exhibit inconsistency speeds, with the maximum latency equivalent to that of the single-frame deep model. In contrast to these, our approach does not use keyframe features, and substitutes optical-flow with an attention propagation module, which we show improves both efficiency and robustness to motion.

%% file: method.tex
\section{Temporally Distributed Network}
In this section, we describe the architecture of a Temporally Distributed Network (TDNet), with an overview in Fig~\ref{fig2}.
In Sec.~\ref{sec_dn} we introduce the main idea of distributing sub-networks to extract feature groups from different temporal frames. In Sec~\ref{sec_tda}, we present our attention propagation module designed for effective aggregation of spatially misaligned feature groups.

\subsection{Distributed Networks}
\label{sec_dn}

\begin{figure}
\centering
\includegraphics[width=1\linewidth]{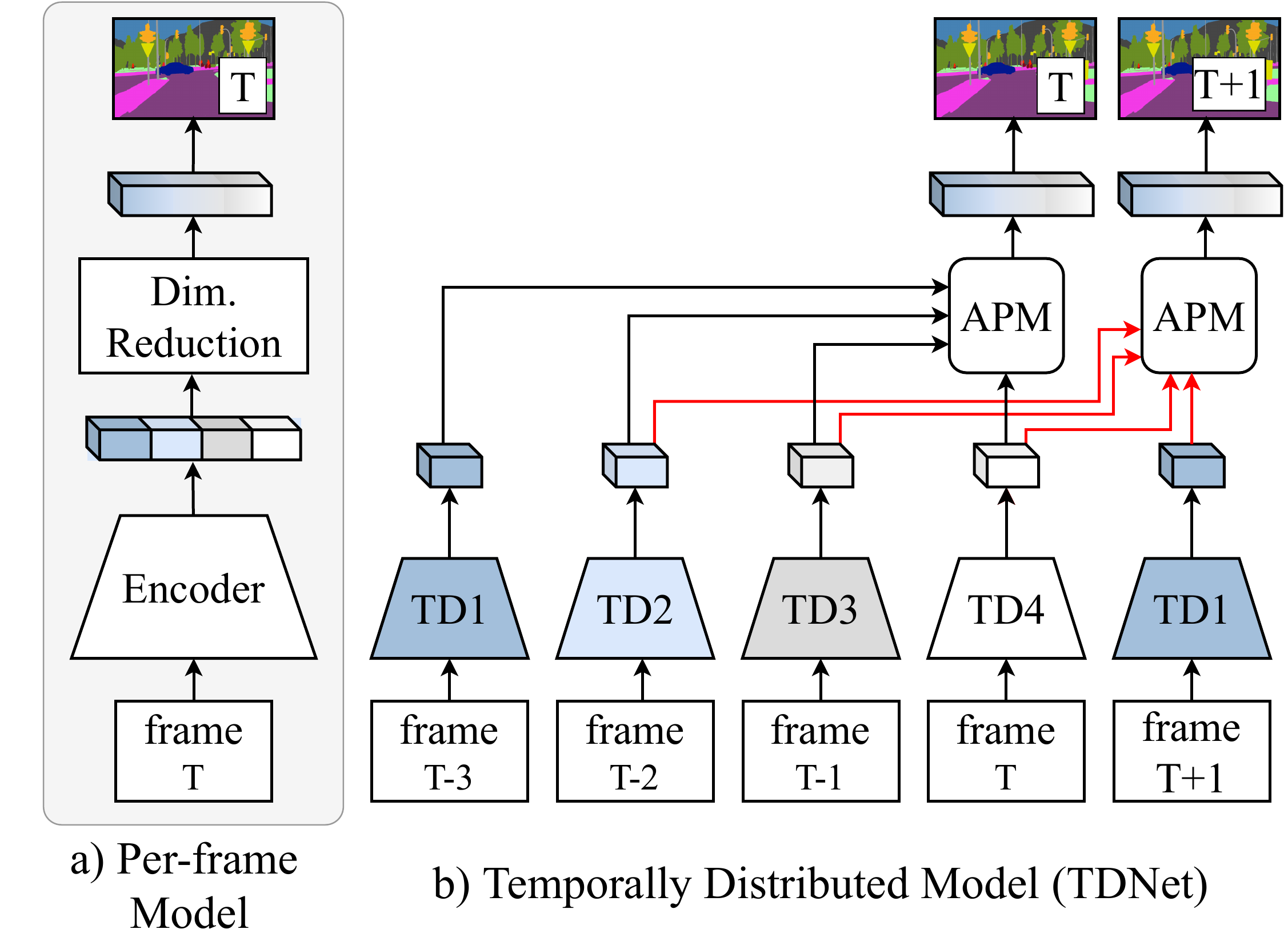}
\caption{As opposed to applying a single deep model to segment each frame independently (a), in TDNet (b) we distribute feature extraction evenly across sequential frames to reduce redundant computation, and then aggregate them using the Attention Propagation Module (APM), to achieve strong features for accurate segmentation.}
\label{fig2}
\vspace{-0.3cm}
\end{figure}

Inspired by the recent success of \textit{Group Convolution}~\cite{krizhevsky2012imagenet,ioannou2017deep} which show that adopting separate convolutional paths can increase a model's effectiveness by enhancing the sparsity of filter relationships, we propose to divide features from a deep neural network into a group of sub-features and approximate them using a set of shallow sub-networks each of which only covers a subspace of the original model's feature representation. 

In addition, we observe that the full feature map is large, and dimension reduction (Fig~\ref{fig2}(a)) is costly.
In PSPNet50~\cite{zhao2017pyramid}, the feature map has 4096 channels and dimension reduction takes about a third of the total computation. 
To further improve efficiency, based on \textit{block matrix multiplication}~\cite{eves1980elementary}, we convert the convolutional layer for dimension reduction to the summation of series of convolution operations at the subspace level, which enables us to distribute these subspace-level convolution operations to their respective subnetworks. 
As a result, the output of the dimension reduction layers is recomposed simply by addition, before being used in the prediction head of the network.
Keeping a similar total model size to the original deep model, we show that aggregating multiple shallow network paths can have a similarly strong representational power as the original deep model~\cite{wu2019wider,Zagoruyko2016WRN,szegedy2017inception,veit2016residual}.

In the context of single image segmentation, the advantage of such an approach is that it allows for faster computation by extracting feature paths \emph{in parallel} on multiple devices. 
However, in the context of segmenting video sequences, we can take advantage of their inherent temporal continuity and distribute the computation along the \emph{temporal} dimension. 
We apply this distributed feature extraction method to video by applying the sub-networks to sequential frames, and refer to the new architecture as Temporally Distributed Network (TDNet).
As shown in Fig~\ref{fig2}(b), TDNet avoids redundant sub-features computation by \emph{reusing} the sub-feature groups computed at previous time steps.
The full feature representation at each frame is then produced by aggregating previously computed feature groups with the current one. 

\subsection{Feature Aggregation}

\begin{figure*}[t!]
\centering
\includegraphics[width=1\linewidth]{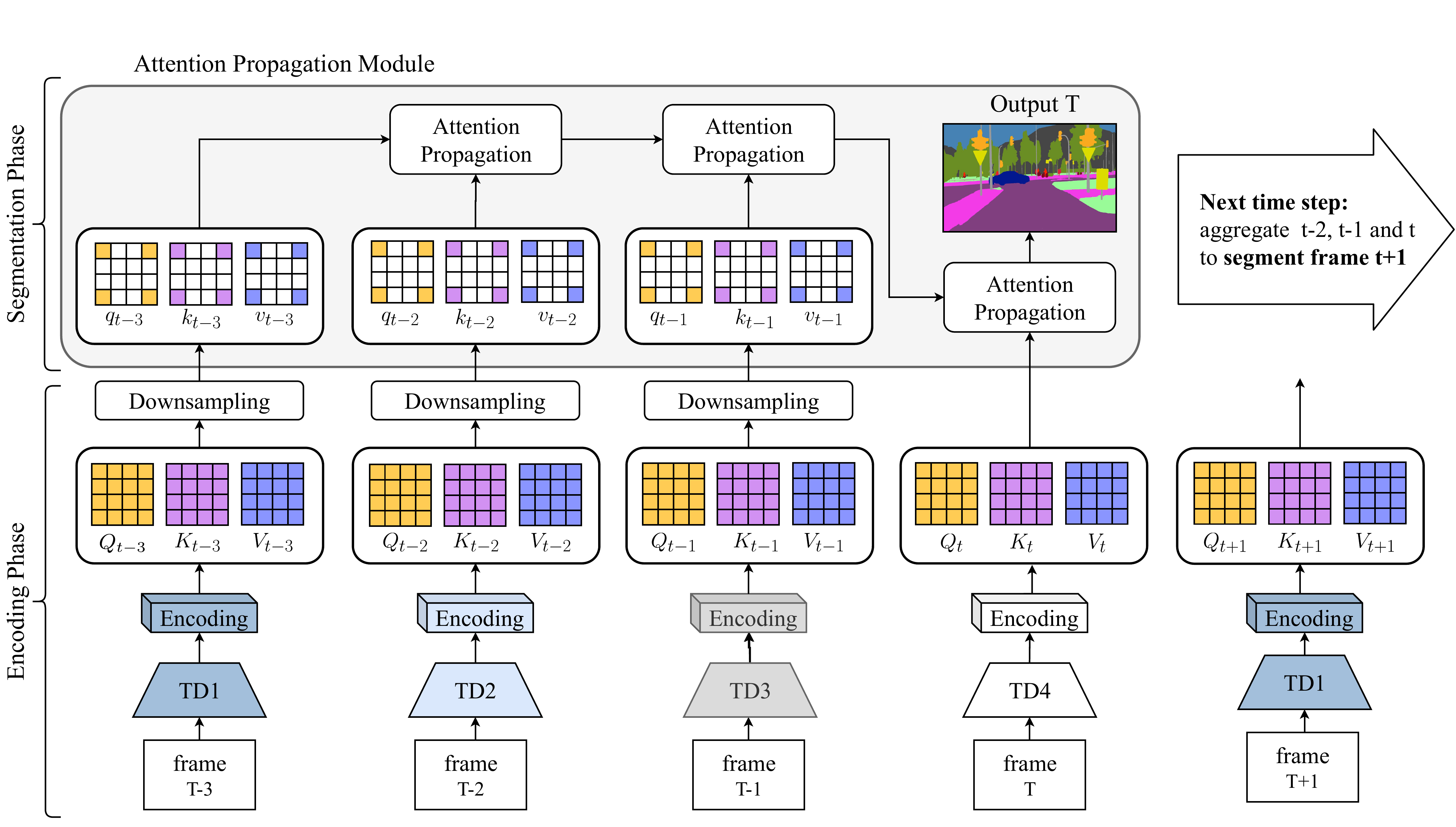}
\caption{\small{Illustration of TDNet with four sub-networks. Since we circularly distribute sub-networks over sequential frames, any four-frame temporal window will cover a full set of the sub-networks. In order to segment frame \textit{t}, we apply the attention propagation module to propagate and merge sub-feature maps previously extracted from (\textit{t-3}, \textit{t-2}, \textit{t-1}) with the sub-feature map from \textit{t}. For the next frame \textit{t+1}, a full feature representation is aggregated by similarly reusing the sub-features extract at frames (\textit{t-2}, \textit{t-1}, \textit{t}). 
}}
\label{fig3}
\vspace{-0.3cm}
\end{figure*}
\label{sec_tda}

A big challenge of aggregating feature groups extracted at different time steps is the spatial misalignment caused by motion between frames.
Optical flow-based warping is a popular tool to correct for such changes~\cite{jain2019accel,zhu2017deep,gadde2017semantic,nilsson2018semantic}, but it is expensive to compute, prone to errors, and restricted to a single match per pixel. 
To tackle such challenges, we propose an Attention Propagation Module (APM), which is based on the non-local attention mechanism~\cite{vaswani2017attention,wang2018non,zhu2019empirical}, but extended to deal with spatio-temporal variations for the video semantic segmentation task.
We now define how we integrate the APM into TDNet. 

As shown in Fig.~\ref{fig3}, TDNet is composed of two phases, the \textit{Encoding Phase} and \textit{Segmentation Phase}. 
The encoding phase extracts alternating sub-feature maps over time. 
Rather than just generating the \textbf{\textit{Value}} feature maps which contain the path-specific sub-feature groups, we also let the sub-networks produce \textbf{\textit{Query}} and \textbf{\textit{Key}} maps for building correlations between pixels across frames. 
Formally, the feature path-$i$ produces a sub-feature map $X_i\in \mathcal{R}^{C\times H\times W}$. 
Then, as in prior work~\cite{vaswani2017attention}, the corresponding encoding module ``Encoding-$i$'' converts $X_i$ into a value map $V_i\in \mathcal{R}^{C\times H\times W}$, as well as lower dimensional query and key maps $Q_i\in \mathcal{R}^{\frac{C}{8}\times H\times W}$, $K_i\in \mathcal{R}^{\frac{C}{8}\times H\times W}$ with three 1$\times$1 convolutional layers.

In the segmentation phase, the goal is to produce segmentation results based on the full features recomposed from the outputs of sub-networks from previous frames. 
Assuming we have $m$ ($m$=4 in Fig.~\ref{fig3}) independent feature paths derived from video frames, and would like to build a full feature representation for frame $t$ by combining the outputs of the previous $m$-1 frames with the current frame.
We achieve this with spatio-temporal attention~\cite{wang2018non,oh2019video}, where we independently compute the \textit{\textbf{Affinity}} between pixels of the current frame $t$ and the previous $m$-1 frames. 
\begin{equation}
\label{eq1}
\begin{split}
\textbf{\textit{Aff}}_{p}= Softmax(\frac{Q_t K_p^{\top}}{\sqrt{d_k}})
\end{split}
\end{equation}
\noindent where $\textit{p}$ indicates a previous frame and $d_k$ is the dimension of the \textbf{\textit{Query}} and \textbf{\textit{Key}}. 
Then, the sub-feature maps at the current frame and previous $m$-1 frames are merged as,
\begin{equation}
\label{eq2}
\begin{split}
V'_{t}= V_t + \sum_{p=t-m+1}^{t-1}\phi(\textbf{\textit{Aff}}_{p} V_p)
\end{split}
\end{equation}
With this attention mechanism, we effectively capture the non-local correlation between pixels across frames, with time complexity of $\mathcal{O}((m-1)d_k H^2W^2)$ for the affinity in Eq.~\ref{eq1}. 
However, features for semantic segmentation are high resolution and Eq~\ref{eq2} incurs a high computation cost.
To improve efficiency, we downsample the attention maps and propagate them over time. 
\paragraph{Attention Downsampling.} We adopt a simple yet effective strategy, which is to downsample the reference data as indicated by the ``Downsampling'' module in Fig.~\ref{fig3}. 
Formally, when segmenting a frame $T$, we apply a spatial pooling operation $\gamma_n (\cdot)$ with stride $n$ to the previous $m$-1 frames' Query, Key, and Value maps,
\begin{equation}
\label{eq3}
\begin{split}
q_i =  \gamma_n(Q_i), \quad k_i =  \gamma_n(K_i), \quad v_i =  \gamma_n(V_i)
\end{split}
\end{equation}
With these downsampled maps, the complexity for Eq.~\ref{eq2} decreases to  $\mathcal{O}(\frac{(m-1)d_k H^2W^2)}{n^2})$. 
We conduct experiments and find that $n$=$4$ works well to preserve necessary spatial information while greatly decreasing the computational cost (see Sec~\ref{sec:downsampling}).

\paragraph{Attention Propagation.} 
Next, we propose a propagation approach, where instead of computing the attention between the current frame and all previous ones, we restrict computation to neighboring frames, and propagate it through the window. 
This allows us not only to reduce the number of attention maps we have to compute, but also to also restrict attention computation to subsequent frames, where motion is smaller. 
Given a time window composed of frames from $t-m+1$ to $t$ together their respective downsampled Query, Key, and Value maps, then for an intermediate frame $p\in(t-m+1,t)$, the attention is propagated as,
\begin{equation}
\label{eq4}
\begin{split}
v'_{p} = \phi\left(Softmax(\frac{q_p  k_{p-1}^{\top}}{\sqrt{d_k}}) v'_{p-1} \right) +v_p 
\end{split}
\end{equation}
where $v'_{t-m+1}=\gamma_n(V_{t-m+1})$, $q$, $k$, and $v$ are the downsampled maps as in Eq.~\ref{eq3},  $d_k$ is the number of dimensions for Query and Key, and $\phi_p$ is a 1$\times$1 convolutional layer.
The final feature representation at frame $t$ is then computed~as,
\begin{equation}
\label{eq5}
\begin{split}
V'_{t}= \phi\left(Softmax(\frac{Q_t k_{t-1}^{\top}}{\sqrt{d_k}}) v'_{t-1}\right) + V_{t}
\end{split}
\end{equation}
and segmentation maps are generated by: $S_{m}= \pi_m(V'_m)$, where $\pi_m$ is the final prediction layer associated with subnetwork $m$.

With this proposed framework, the time complexity is reduced to $\mathcal{O}( \frac{(m-2)\cdot d_k H^2W^2)}{n^4} + \frac{d_k H^2W^2)}{n^2}) \approx \mathcal{O}( \frac{d_k H^2W^2)}{n^2})$. 
Since the attention is extracted from neighboring frames only, the resulting feature are also more robust to scene motion. 
We notice that recent work~\cite{Zhu_2019_ICCV} also adopt pooling operation to achieve efficient attention models, but this is in the context of image semantic segmentation, while our model extends this strategy to deal with video data.

\section{Grouped Knowledge Distillation} 
\label{sec_gkd}

During training, we further enhance the complementarity of sub-feature maps in the full feature space by introducing a knowledge distillation~\cite{hinton2015distilling} strategy, using a strong deep model designed for single images as the teacher network. 
In addition to transferring knowledge in the full-feature space~\cite{hinton2015distilling,Liu_2019_CVPR,He_2019_CVPR}, we propose a grouped knowledge distillation loss to further transfer knowledge at the subspace level in order to make the information extracted from different paths more complementary to one another.

\begin{figure}
\centering
\includegraphics[width=0.85\linewidth]{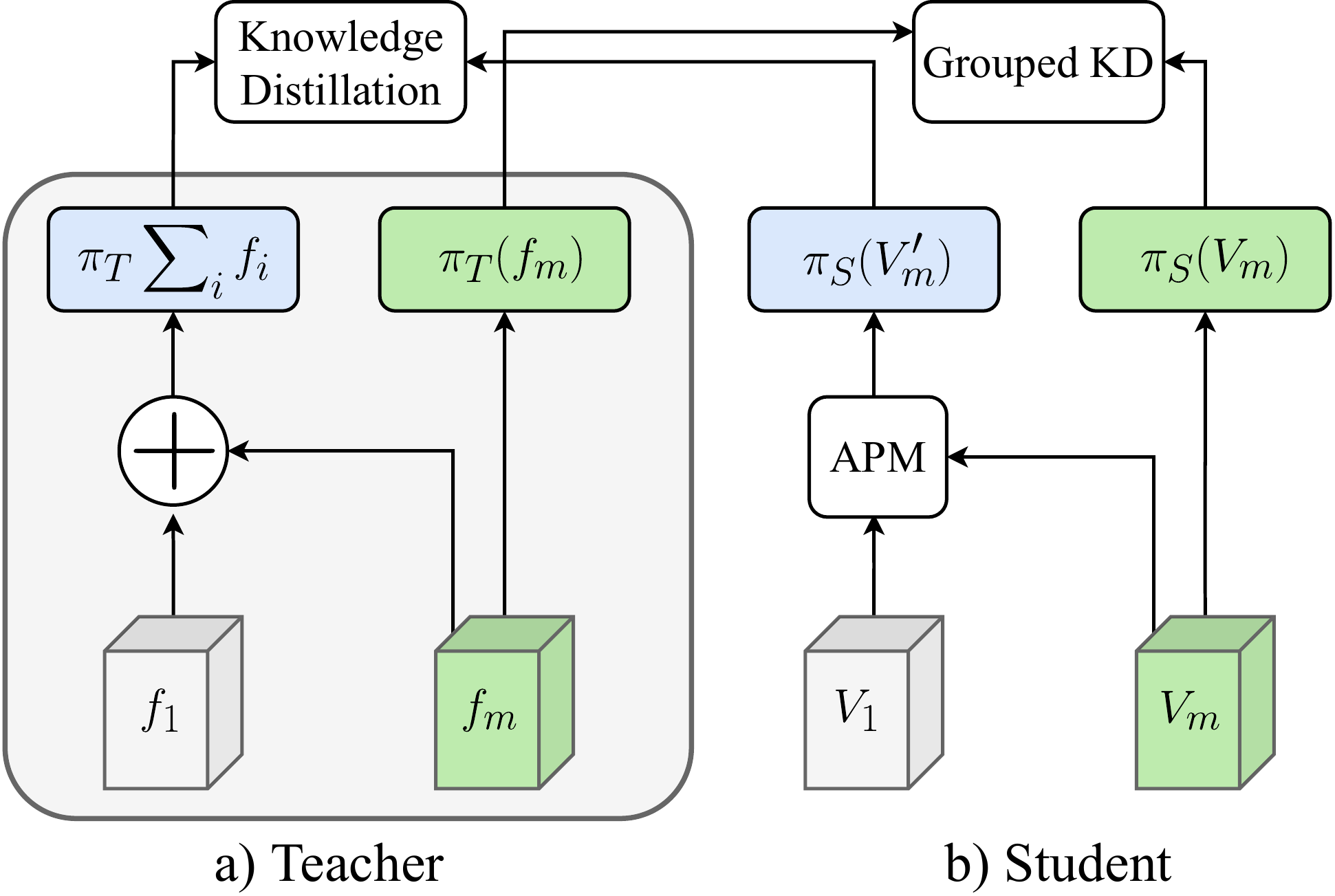}
\vspace{-0.1cm}
\caption{\small{The knowledge distillation. In the ``Overall KD'', we align the  full outputs between the teacher model (e.g. PSPNet101) and the student model (e.g. out TDNet). In the ``Grouped KD'', we match the outputs  based on only one sub-network to the teacher model's output conditioned on the respective feature subspace.}}
\label{fig4}
\vspace{-0.5cm}
\end{figure}

The idea of a grouped distillation loss is illustrated in Fig.~\ref{fig4}. 
We take a deep baseline model like PSPNet101 as the teacher, and take our TDNet with $m$ sub-networks as the student network. 
The goal is to not only align the output distributions at the whole-model level, but also at a subfeature group level. 
Based on \textit{block matrix multiplication}~\cite{eves1980elementary}, we evenly separate the teacher model's feature reduction layer into $m$ independent sub-convolution groups, which output a set of sub-feature groups $\{f_i|i=1,...,m\}$. 
Thus, the original segmentation result is $\pi_T(\sum f)$, and the contribution of the $i$-th feature group is $\pi_T(f_i)$, given  $\pi_T(\cdot)$ being the teacher model's segmentation layer. 
In TDNet, the target frame's Value map $V_m$ is combined with propagated previous information to be $V'_m$, thus the full model output is $\pi_S(V'_m)$ and the $m$-th feature path's contribution is $\pi_S(V_m)$, given $\pi_S(\cdot)$ is the final segmentation layers. Based on these, our final loss function is,
\begin{align}
Loss= &CE(\pi_S(V'_i, gt)) + \alpha\cdot KL(\pi_S(V'_i)||\pi_T(\sum f))\nonumber\\
      &+\beta\cdot KL(\pi_S(V_i)||\pi_T(f_i))
\label{eq7}
\end{align}
\noindent where $CE$ is the cross entropy loss, and $KL$ means the KL-divergence. The first term is the supervised training with ground truth. The second term distills knowledge at the whole-model level. The third term transfers knowledge at feature group level. We set $\alpha$ and $\beta$ to be 0.5 in our paper.

%% file: experiments.tex
\section{Experiments}

We evaluate our method on Cityscapes~\cite{cordts2016cityscapes} and Camvid~\cite{brostow2008segmentation} for street views, and NYUDv2~\cite{Silberman12nyyd} for indoor scenes. 
On all of these datasets, our method achieves state-of-the-art accuracy with a much faster speed and lower and evenly distributed latency.

\subsection{Setup and Implementation}

\paragraph{Datasets \& Evaluation Metrics.}
\textit{Cityscapes}~\cite{cordts2016cityscapes} contains 2,975/500/1,525 snippets for training/validation/testing. The 20$^{th}$ frame of each snippet is annotated with 19 classes for semantic segmentation.
\textit{Camvid}~\cite{brostow2008segmentation} consists of 4 videos with 11-class pixelwise annotations at 1Hz. 
The annotated frames are grouped into 467/100/233 for training/validation/testing. 
\textit{NYUDv2}~\cite{Silberman12nyyd} contains 518 indoor videos with 795 training frames and 654 testing frames being rectified and annotated with 40-class semantic labels. 
Based on these labeled frames, we create rectified video snippets from the raw Kinetic videos, which we will release for testing. 
Following the practice in previous works~\cite{he2017std2p,li2018low,jin2017video,gadde2017semantic}, we evaluate mean Intersection-over-Union (mIoU) on Cityscapes, and mean accuracy and mIoU on Camvid and NYUDv2.

\paragraph{Models \& Baselines.} We demonstrate the effectiveness of TDNet on different backbones. 
We select two state-of-the-art image segmentation models for our experiments: PSPNet~\cite{zhao2017pyramid}, and BiseNet$^*$ ~\cite{yu2018bisenet}. 
The latter is a modified/improved version of~\cite{yu2018bisenet} with the \textit{Spatial Path} being replaced with the output of ResBlock-2, which we found to have higher efficiency and better training convergence. 
We extend these image models with temporally distributed framework to boost the performance, yielding the models:
\\ 
\noindent \textbf{TD$^2$-PSP50}, \textbf{TD$^4$-PSP18}: the former consists of two PSPNet-50~\cite{zhao2017pyramid} backbones with halved output channels as sub-networks, whereas TD$^4$-PSP18 is  made of four PSPNet-18 sub-networks. The model capacity of the temporally distributed models is comparable to the image segmentation network they are based on (PSPNet-101).
\\ 
\noindent\textbf{TD$^2$-Bise34}, \textbf{TD$^4$-Bise18}. 
Similarly, we build TD$^2$-Bise34 with two BiseNet$^*$-34 as sub-networks, and TD$^4$-Bise18 with four BiseNet$^*$-18 as sub-networks for the real-time applications. Like in PSPNet case, the model capacity of the temporally distributed networks is comparable to the BiseNet$^*$-101.

\paragraph{Speed Measurement \& Comparison.} All testing experiments are conducted with a batch-size of one on a single Titan Xp in the Pytorch framework. 
We found that previous methods are implemented with different deep-learning frameworks and evaluated on different types of devices, so for consistent comparisons, we report the speed/latency for these previous methods based on benchmark-based conversions\footnote{http://goo.gl/N6ukTz/, http://goo.gl/BaopYQ/} and our reimplementations. 

\paragraph{Training \& Testing Details.} 
Both our models and baselines are initialized with Imagenet~\cite{deng2009imagenet} pretrained parameters and then trained to convergence to achieve the best performance.
To train TDNet with $m$ subnetworks, each training sample is composed of $m$ consecutive frames and the supervision is the ground truth from the last one. 
We perform random cropping, random scaling and flipping for data augmentation.
Networks are trained by stochastic gradient descent with momentum 0.9 and weight decay 5e-4 for 80k iterations. 
The learning rate is initialized as 0.01 and decayed by $(1-\frac{iter}{max-iter})^{0.9}$. 
During testing, we resize the output to the input's original resolution for evaluation. 
On datasets like Cityscapes and NYUDv2 which have temporally sparse annotations, we compute the accuracy for all possible orders of sub-networks and average them as final results.
We found that different orders of sub-networks achieve very similar mIoU values, which indicates that TDNet is stable with respect to sub-feature paths (see supplementary materials).

\subsection{Results}
\begin{table}
\centering
\begin{tabular}{ |c|cc|c|c|} 
\hline
Method             &\multicolumn{2}{c|}{mIoU($\%$)}       &Speed &Max Latency \\
                     &\textit{val} &\textit{test}    &(ms/f)   &(ms) \\
 \hline 
 \hline 
 \small{CLK~\cite{shelhamer2016clockwork} }             &64.4 &-    &158 &198\\ 
 \small{DFF~\cite{zhu2017deep}}                         &69.2 &-    &156 &575\\ 
 \small{GRFP(5)~\cite{nilsson2018semantic}}             &73.6 &72.9 &255 &255\\ 
 \small{LVS-LLS~\cite{li2018low}}                       &75.9 &-    &\underline{119} &\underline{119}\\
 \small{PEARL~\cite{jin2017video}}                      &76.5 &75.2 &800 &800 \\ 
 \small{LVS~\cite{li2018low}}                           &\underline{76.8} &-    &171 &380\\  
 \hline
 \small{PSPNet18~\cite{zhao2017pyramid}}                  &75.5 &-    &91  &91    \\
 \small{PSPNet50~\cite{zhao2017pyramid}}                  &78.1 &-    &238 &238   \\ 
 \small{PSPNet101~\cite{zhao2017pyramid}}                 &79.7 &79.2 &360 &360   \\ 
 \hline 
 \small{\textbf{TD$^4$-PSP18}}                             &\underline{76.8} &- &\textbf{85}  &\textbf{85}    \\ 
 \small{\textbf{TD$^2$-PSP50}}  &\textbf{79.9} &79.4 &178 &178   \\ 
 \hline 
\end{tabular}
\vspace{0.1cm}
\caption{\small{Evaluation on the Cityscapes dataset. The ``Speed'' and ``Max Latency'' represent the average and maximum per-frame time cost respectively.}}
\label{tab1}
\vspace{-0.5cm}
\end{table}

\begin{figure*}[t]
\centering
\includegraphics[height=6.4cm]{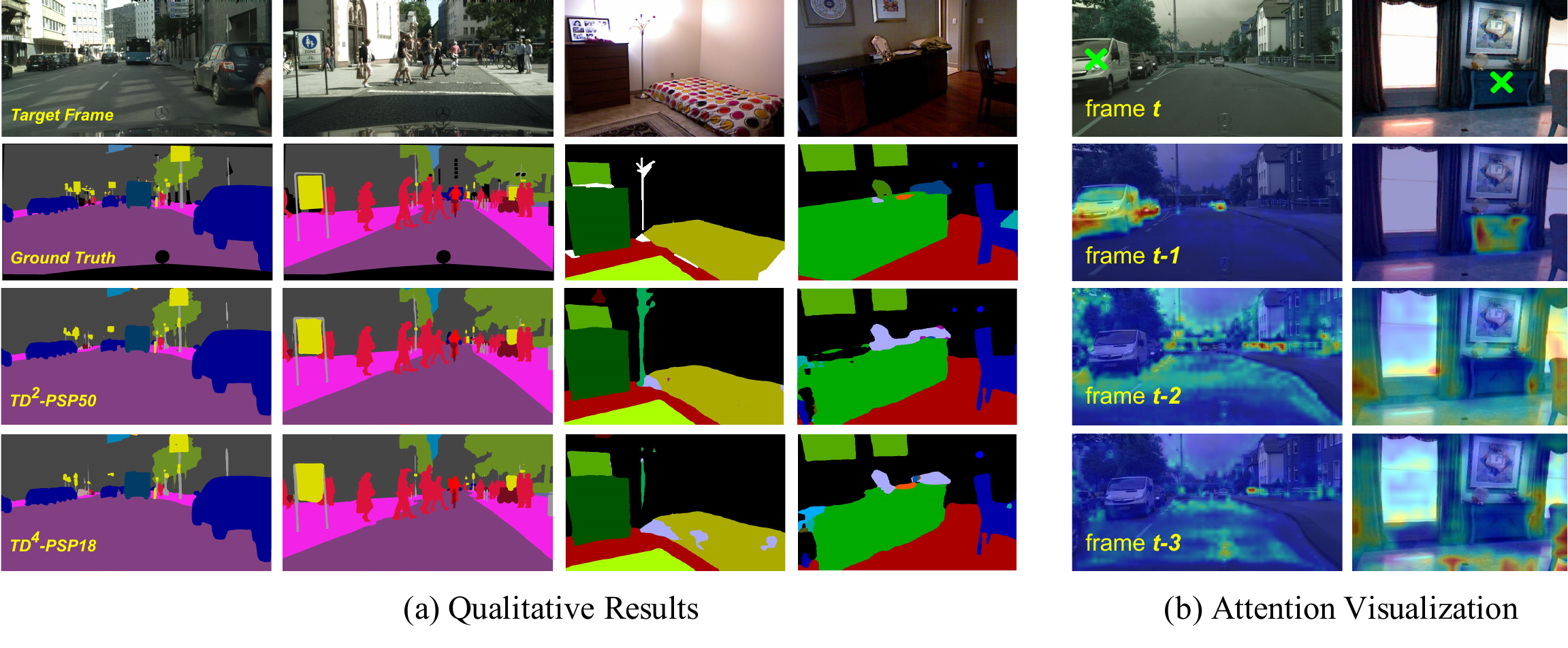}
\vspace{-0.3cm}
\caption{\small{ Qualitative results of our method on Cityscapes and NYUD-v2 (a), and a visualization of the attention map in our attentive propagation network (b). Given a pixel in frame \textit{t} (denoted as a green cross), we back-propagate the correlation scores with the affinity matrices, and then visualize the normalized soft weights as heat map over the other frames in the window.}
}
\label{fig5}
\vspace{-0.4cm}
\end{figure*}

\paragraph{Cityscapes Dataset.}
We compare our method with the recent state-of-the-art models for semantic video segmentation in Table~\ref{tab1}. Compared with LVS~\cite{li2018low}, TD$^4$-PSP18, achieves similar performance with only a half the average time cost, and TD$^2$-PSP50 further improves accuracy by 3 percent in terms of mIoU.
Unlike keyframe-based methods like LVS~\cite{li2018low},  ClockNet~\cite{shelhamer2016clockwork}, DFF~\cite{zhu2017deep} that have fluctuating latency between keyframes and non-key frames (e.g. 575ms v.s. 156ms for DFF~\cite{zhu2017deep}), our method runs with a balanced computation load over time.
With a similar total number of parameters as PSPNet101~\cite{zhao2017pyramid}, TD$^2$-PSP50 reduces the per-frame time cost by half from 360ms to 178ms while improving accuracy.
The sub-networks in  TD$^2$-PSP50 are adapted from PSPNet50, so we also compare their performance, and can see that TD$^2$-PSP50 outperforms PSPNet50 by $1.8\%$ mIoU with a faster average latency. 
As shown in the last row, TD$^4$-PSP18 can further reduce the latency to a quarter, but due to the shallow sub-networks (based on a PSPNet18 model), the performance drops comparing to PSPNet101.
However, it still achieves state-of-the-art accuracy and outperforms previous methods by a large gap in terms of latency.
Some qualitative results are shown in Fig.~\ref{fig5}(a)

\begin{table}
\centering
\begin{tabular}{ |c|cc|c|} 
\hline
{Method}       &\multicolumn{2}{c|}{{mIoU($\%$)}}     &{Speed (ms/f)}  \\
                     &{~~\textit{val}~~} &{~~\textit{test}~~}     &{}     \\
 \hline 
 \hline 
 \small{DVSNet~\cite{xu2018dynamic} }                       &63.2 &-    &33 \\ 
 \small{ICNet~\cite{zhao2018icnet} }                       &67.7 &69.5    &\textbf{20} \\ 
 \small{LadderNet~\cite{kreso2017ladder} }                 &72.8 &-       &33 \\ 
 \small{SwiftNet~\cite{orsic2019defense} }                 &75.4 &-    &23 \\ 
 \hline  
 \small{BiseNet$^*$18~\cite{yu2018bisenet}}          &73.8 &73.5  &20\\  
 \small{BiseNet$^*$34~\cite{yu2018bisenet}}          &76.0 &-  &27\\ 
 \small{BiseNet$^*$101~\cite{yu2018bisenet}}          &76.5 &-  &72\\ 
 \hline 
 \small{\textbf{TD$^4$-Bise18}}                   &75.0 &74.9  &21\\ 
 \small{\textbf{TD$^2$-Bise34}}                   &\textbf{76.4} &-       & 26 \\ 
 \hline
\end{tabular}
\vspace{0.1cm}
\caption{\small{Evaluation of high-efficiency approaches on the Cityscapes dataset.}}
\label{tab2}
\vspace{-0.5cm}
\end{table}

To validate our method's effectiveness for more realistic tasks, we evaluate our real-time models TD$^2$-Bise34 and TD$^4$-Bise18 (Table~\ref{tab2}).
As we can see, TD$^2$-Bise34 outperforms all the previous real-time methods like ICNet~\cite{zhao2018icnet}, LadderNet~\cite{kreso2017ladder}, and SwiftNet~\cite{orsic2019defense} by a large gap, at a comparable, real-time speed.
With a similar total model size to BiseNet$^*$101, TD$^2$-Bise34 achieves better performance while being roughly three times faster.
TD$^4$-Bise18 drops the accuracy but further improves the speed to nearly 50 FPS. 
Both TD$^2$-Bise34 and TD$^4$-Bise18 improve over their single path baselines at a similar time cost, which validates the effectiveness of our TDNet for real-time tasks.

\paragraph{Camvid Dataset.}
We also report the evaluation of Camvid dataset in Table~\ref{tab3}. We can see that TD$^2$-PSP50 outperforms the previous state-of-the-art method Netwarp~\cite{gadde2017semantic} by about 9\% mIoU while being roughly four times faster. 
Comparing to the PSPNet101 baselines with a similar model capacity, TD$^2$-PSP50 reduces about half of the computation cost with comparable accuracy. 
The four-path version further reduces the latency by half but also decreases the accuracy. 
This again shows that a proper depth is necessary for feature path, although even so, TD$^4$-PSP18 still outperforms previous methods with a large gap both in terms of mIoU and speed.

\begin{table}
\centering
\small
\begin{tabular}{ |c|c|c|c|} 
 \hline
 \small{Method}      &\small{mIoU($\%$)} &\small{Mean Acc.($\%$)} &\small{Speed(ms/f)}\\ 
 \hline 
 \hline 
 \footnotesize{LVS~\cite{li2018low}}               &-       &82.9           &84\\ 
 \footnotesize{PEARL~\cite{jin2017video} }         &-       &\underline{83.2}          &300 \\ 
 \footnotesize{GRFP(5)~\cite{nilsson2018semantic}} &66.1    &-              &230\\ 
 \footnotesize{ACCEL~\cite{jain2019accel}}         &66.7    &-              &132\\ 
 \footnotesize{Netwarp~\cite{gadde2017semantic} }  &67.1    &-              &363 \\ 
 \hline 
 \footnotesize{PSPNet18~\cite{zhao2017pyramid}}      &71.0       &78.7          &40\\ 
 \footnotesize{PSPNet50~\cite{zhao2017pyramid}}      &74.7       &81.5          &100\\ 
 \footnotesize{PSPNet101~\cite{zhao2017pyramid}}      &76.2       &83.6          &175\\ 
 \hline 
 \footnotesize{\textbf{TD$^4$-PSP18}}               &\underline{72.6}  &80.2          &\textbf{40}\\ 
 \footnotesize{\textbf{TD$^2$-PSP50}}               &\textbf{76.0}    &\textbf{83.4}          &\underline{90}\\ 
 \hline
\end{tabular}
\vspace{0.1cm}
\caption{\small{Evaluation on the Camvid dataset.}}
\label{tab3}
\vspace{-0.1cm}
\end{table}

\paragraph{NYUDv2 Dataset.}
To show that our method is not limited to street-view like scenes, we also reorganize the indoor NYUDepth-v2 dataset to make it suitable for semantic video segmentation task.  
As most previous methods for video semantic segmentation do not evaluate on this dataset, we only find one related work to compare against; STD2P~\cite{he2017std2p}. 
As shown in Table~\ref{tab4}, TD$^2$-PSP50 outperforms STD2P in terms of both accuracy and speed. 
TD$^4$-PSP18 achieves a worse accuracy but is more than 5$\times$ faster. 
TD$^2$-PSP50 again successfully halves the latency but keeps the accuracy of the baseline PSPNet101, and also achieves about 1.6$\%$ improvement in mIoU comparing to PSPNet18 without increasing the latency.

\begin{table}
\centering
\small
\begin{tabular}{ |c|c|c|c|} 
 \hline
 \small{Method}      &\small{mIoU($\%$)} &\small{Mean Acc.($\%$)} &\small{Speed(ms/f)}\\ 
 \hline 
 \hline 
 \footnotesize{STD2P~\cite{he2017std2p} }          &\underline{40.1}    &\underline{53.8}               &$>$100\\ 
 \hline 
 \footnotesize{FCN~\cite{long2015fully} }          &34.0                &46.1               &56\\ 
 \footnotesize{DeepLab~\cite{chen2017deeplab} }    &39.4                &49.6   &78\\ 
 \footnotesize{PSPNet18~\cite{zhao2017pyramid}}      &35.9              &46.9               &19\\ 
 \footnotesize{PSPNet50~\cite{zhao2017pyramid}}      &41.8              &52.8               &47\\
 \footnotesize{PSPNet101~\cite{zhao2017pyramid}}     &43.2              &55.0               &72\\
 \hline 
 \footnotesize{\textbf{TD$^4$-PSP18}}                &37.4              &48.1               &\textbf{19}\\ 
 \footnotesize{\textbf{TD$^2$-PSP50}}                &\textbf{43.5}     &\textbf{55.2}      &\underline{35}\\ 
 \hline
\end{tabular}
\vspace{0.1cm}
\caption{\small{Evaluation on the NYUDepth dataset.}}
\label{tab4}
\vspace{-0.2cm}
\end{table}

\subsection{Method Analysis}
\begin{table}[t]
\small
\centering
\begin{tabular}{|cc|p{1.3cm} p{1.3cm}|} 
\hline
\footnotesize{Overall-KD}  &\footnotesize{Grouped-KD}&\footnotesize{Cityscapes}  &\footnotesize{NYUDv2}\\ 
 \hline  
 \hline 
  &  &76.4  &36.2\\ 
\footnotesize{$\checkmark$}  &  & 76.5 \scriptsize{(+0.1)} &36.7 \scriptsize{(+0.5)} \\
\footnotesize{$\checkmark$}  &\footnotesize{$\checkmark$}  &\textbf{76.8 \scriptsize{(+0.4)}}   &\textbf{37.4 \scriptsize{(+1.2)}} \\ 
 \hline
\end{tabular}
\vspace{0.1cm}
\caption{\small{The mIoU ($\%$) for different components in our knowledge distillation loss (Eq.~\ref{eq7}) for TD$^4$-PSP18.}}
\label{tab5}
\vspace{-0.5cm}
\end{table}

\paragraph{Grouped Knowledge Distillation.} The knowledge distillation based training loss (Eq.~\ref{eq7}) consistently helps to improve performance on the three datasets. 
In order to investigate the effect of different components in the loss, we train TD$^4$-PSP18 with different settings and show the results in Table~\ref{tab5}. 
The overall knowledge distillation~\cite{hinton2015distilling} works by providing extra information about intra-class similarity and inter-class diversity. 
Thereby, it is less effective to improve a fully trained base model on Cityscapes due to the highly-structured contents and relatively fewer categories. 
However, when combined with our grouped knowledge distillation, the performance can be still boosted with nearly a half percent in terms of mIoU.
This shows the effectiveness of our grouped knowledge distillation to provide extra regularization.
On the NYUD-v2 dataset which contains more diverse scenes and more categories, our method achieves significant improvements with an 1.2$\%$ absolute improvement in mIoU. 

\begin{figure}[h]
\centering
\includegraphics[height=4cm]{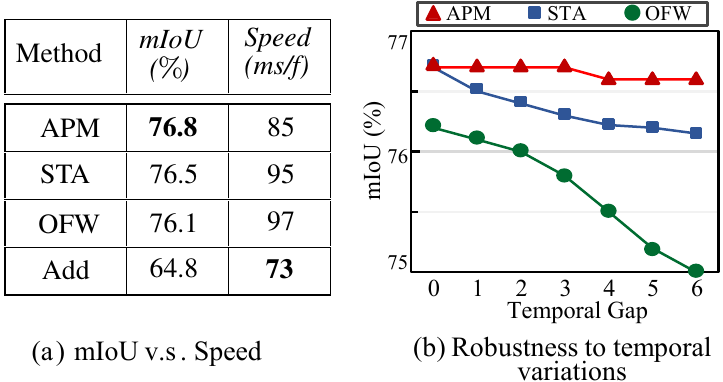}
\caption{\small{TD$^4$-PSP18 with different temporal aggregation methods on Cityscapes dataset. ``APM'' denotes our attention propagation module. ``STA'' represents spatio-temporal attention~\cite{wang2018non,oh2019video}. ``OFW'' is the optical-flow~\cite{dosovitskiy2015flownet} based fusion. ``Add'' means simply adding feature maps.} 
}
\label{fig6}
\vspace{-0.2cm}
\end{figure}

\paragraph{Attention Propagation Module.} 
Here, we compare our attention propagation module (APM) with other aggregation methods such as: no motion compensation, e.g., just adding feature groups (Add), optical-flow based warping (OFW) and the vanilla Spatio-Temporal Attention (STA) mechanism~\cite{wang2018non,oh2019video}. 
As shown in Fig.~\ref{fig6}(a), without considering the spatial misalignment (Add) leads to the worst accuracy. 
Our APM outperforms OFW and STA in both accuracy and latency.
In Fig.~\ref{fig6}(b), we evaluate our method's robustness to motion between frames by varying the temporal step in input frames sampling.
As shown in the figure, APM shows the best robustness, even with a sampling gap of 6 frames where flow based methods fail, our APM drops very slightly in contrast to other methods.

\paragraph{Attention Downsampling.} 
\label{sec:downsampling}
In the downsampling operation used to improve the efficiency of computing attention, we apply spatial max pooling with a stride $n$. 
We show the influence of $n$ in Table~\ref{tab6}. 
By increasing $n$ from 1 to 4, the computation is decreased drastically, while the accuracy is fairly stable. 
This indicates that the downsampling strategy is effective in extracting spatial information in a sparse way. 
However, while further increasing $n$ to 32, the accuracy decreases due to the information being too sparse.

\begin{table}[t]
\centering
\resizebox{1.\linewidth}{!}{
\begin{tabular}{|c|c|cccccc|}
\hline
\multicolumn{2}{|c|}{{Model}}  &{$n$=1} &{~2} &{~4} &{~8} &{~16}  &{~32}\\
 \hline  
 \hline  
\multirow{2}*{\footnotesize{TD$^2$-PSP50}}   &\footnotesize{mIoU ($\%$)}  &{80.0}   &{80.0}   &{79.9} &{79.8}   &{79.6}   &{79.1}  \\
                                            &\footnotesize{latency (ms)} &{251}    &{205}    &{178}    &{175}    &{170}    &{169}  \\
 \hline  
\multirow{2}*{\footnotesize{TD$^4$-PSP18}}   &\footnotesize{mIoU ($\%$)}  &{76.9}   &{76.8}   &{76.8} &{76.5}   &{76.1}   &{75.7}  \\
                                            &\footnotesize{latency (ms)} &{268}    &{103}    &{85}     &{81}     &{75}     &{7}5  \\
 \hline  
\multirow{2}*{\footnotesize{TD$^4$-Bise18}}    &\footnotesize{mIoU ($\%$)}  &{75.0}   &{75.0}   &{75.0}   &{74.8}   &{74.7}   &{74.4}  \\
                                            &\footnotesize{latency (ms)} &{140}   &{31}   &{21}   &{19}   &{18}   &{18}  \\
 \hline  
\end{tabular}
}
\vspace{0.05cm}
\caption{Effect of different downsampling stride $n$ on Cityscapes. 
}
\label{tab6}
\end{table}

\begin{table}[t]
\centering
\small
\begin{tabular}{ |c|c|c|c|} 
 \hline
 Framework &Single Path Baseline &Shared & Independent\\ 
 \hline 
 \hline 
 \footnotesize{TD$^2$-PSP50}   &78.2    &78.5 & \textbf{79.9}\\ 
 \footnotesize{TD$^4$-PSP18}   &75.5    &75.7 & \textbf{76.8}\\ 
 \hline
\end{tabular}
\vspace{0.1cm}
\caption{\small{Comparisons on Cityscapes for using a shared sub-network or independent sub-networks. The last column shows the baseline model corresponding to TDNet's sub-network.}}
\label{tab7}
\vspace{-0.6cm}
\end{table}

\paragraph{Shared Subnetworks v.s. Independent Subnetworks.} 
When processing a video, the effectiveness of TDNet may come from two aspects: the enlarged representation capacity by distributed subnetworks and the temporal context information provided by neighboring frames.
In Table~\ref{tab7}, we analyze the contributions of each by using a single subnetwork used for each path, or a group of independent subnetworks. 
As we can see, aggregating features extracted via a shared single subnetwork can improve the performance of image segmentation baseline, and independent sub-networks can further improve mIoU by 1\% without increasing computation cost. 
This shows that TDNet does not only benefit from the temporal context information but is also effectively enlarging the representation capacity by the temporally distributing distinct subnetworks.

\begin{table}[t]
\centering
\begin{tabular}{|cccc|cc|}
\hline
\small{P1}&\small{P2} &\small{P3}&\small{P4} &\small{Cityscapes}&\small{NYUDepth-V2}\\     
\hline   
\hline   
\footnotesize{$\checkmark$}&\footnotesize{$\checkmark$}&\footnotesize{$\checkmark$}&\footnotesize{$\checkmark$}&\footnotesize{\textbf{76.8}}&\footnotesize{\textbf{38.2}}\\
&\footnotesize{$\checkmark$}&\footnotesize{$\checkmark$}&\footnotesize{$\checkmark$}&\footnotesize{76.5}&\footnotesize{38.0}\\
&&\footnotesize{$\checkmark$}&\footnotesize{$\checkmark$}&\footnotesize{76.0}&\footnotesize{37.2}\\
&&&\footnotesize{$\checkmark$}&\footnotesize{74.3}&\footnotesize{34.4}\\
\hline
\end{tabular}
\vspace{2pt}
\caption{Ablation study on  TD$^4$-PSP18 showing how performance decreases with progressively fewer sub-features accumulated.}
\label{tab8}
\vspace{-0.4cm}
\end{table}

\paragraph{Effect of Sub-networks.} 
As shown in the last part, TDNet benefits from enforcing different sub-networks extract complementary feature groups.
Here, we provide detailed ablation studies about the contributions of these sub-networks.
Table~\ref{tab8} shows the analysis for TD$^4$-PSP18, where P4 represents the sub-network at the target frame, and P1$\sim$P3 are the sub-networks applied on the previous frames. 
As we can see, by removing feature paths from the first frame, the accuracy consistently decreases for both datasets, which proves the effectiveness of feature distribution. 
To show how these paths are aggregated, in Fig~\ref{fig5}(b) we visualize the attention maps of the attention propagation module in TD$^4$-PSP18.
As shown in the figure, given a pixel (denoted as green crosses) in the target frame \textit{t}, pixels of the corresponding semantic category in the previous frame \textit{t}-1 are matched. 
However, in the previous frames \textit{t}-2 and \textit{t}-3, \emph{background} pixels are collected. 
It should be noted that in the attention propagation module, there are layers $\phi$ (in Eq.~\ref{eq4} and Eq.~\ref{eq5}) which process the aggregated features. 
Thus frames \textit{t}-2 and \textit{t}-3 provide contextual information, and frames \textit{t}-1 and \textit{t} provide local object information, which are combined together to form strong and robust features for segmentation.

%% file: conclusions.tex
\section{Conclusion}
We presented a novel temporally distributed network for fast semantic video segmentation. 
By computing the feature maps across different frames and merging them with a novel attention propagation module, our method  retains high accuracy while significantly improving the latency of processing video frames. 
We show that using a grouped knowledge distillation loss, further boost the performance. TDNet consistently outperforms previous methods in both accuracy and efficiency.

\noindent \textbf{Acknowledgements.} We thank Kate Saenko for the useful discussions and suggestions. This work was supported in part by DARPA and NSF, and a gift funding from Adobe Research.

%% file: main.bbl
\begin{thebibliography}{10}\itemsep=-1pt

\bibitem{brostow2008segmentation}
Gabriel~J Brostow, Jamie Shotton, Julien Fauqueur, and Roberto Cipolla.
\newblock Segmentation and recognition using structure from motion point
  clouds.
\newblock In {\em ECCV}, 2008.

\bibitem{carreira2018massively}
Joao Carreira, Viorica Patraucean, Laurent Mazare, Andrew Zisserman, and Simon
  Osindero.
\newblock Massively parallel video networks.
\newblock In {\em ECCV}, 2018.

\bibitem{chen2017deeplab}
Liang-Chieh Chen, George Papandreou, Iasonas Kokkinos, Kevin Murphy, and Alan~L
  Yuille.
\newblock Deeplab: Semantic image segmentation with deep convolutional nets,
  atrous convolution, and fully connected crfs.
\newblock {\em IEEE T-PAMI}, 2017.

\bibitem{chen2018encoder}
Liang-Chieh Chen, Yukun Zhu, George Papandreou, Florian Schroff, and Hartwig
  Adam.
\newblock Encoder-decoder with atrous separable convolution for semantic image
  segmentation.
\newblock In {\em ECCV}, 2018.

\bibitem{cordts2016cityscapes}
Marius Cordts, Mohamed Omran, Sebastian Ramos, Timo Rehfeld, Markus Enzweiler,
  Rodrigo Benenson, Uwe Franke, Stefan Roth, and Bernt Schiele.
\newblock The cityscapes dataset for semantic urban scene understanding.
\newblock In {\em CVPR}, 2016.

\bibitem{deng2009imagenet}
Jia Deng, Wei Dong, Richard Socher, Li-Jia Li, Kai Li, and Li Fei-Fei.
\newblock Imagenet: A large-scale hierarchical image database.
\newblock In {\em CVPR}, 2009.

\bibitem{ding2018context}
Henghui Ding, Xudong Jiang, Bing Shuai, Ai Qun~Liu, and Gang Wang.
\newblock Context contrasted feature and gated multi-scale aggregation for
  scene segmentation.
\newblock In {\em CVPR}, 2018.

\bibitem{dosovitskiy2015flownet}
Alexey Dosovitskiy, Philipp Fischer, Eddy Ilg, Philip Hausser, Caner Hazirbas,
  Vladimir Golkov, Patrick Van Der~Smagt, Daniel Cremers, and Thomas Brox.
\newblock Flownet: Learning optical flow with convolutional networks.
\newblock In {\em ICCV}, 2015.

\bibitem{eves1980elementary}
Howard~Whitley Eves.
\newblock {\em Elementary matrix theory}.
\newblock 1980.

\bibitem{gadde2017semantic}
Raghudeep Gadde, Varun Jampani, and Peter~V Gehler.
\newblock Semantic video cnns through representation warping.
\newblock In {\em CVPR}, 2017.

\bibitem{He_2019_ICCV}
Junjun He, Zhongying Deng, and Yu Qiao.
\newblock Dynamic multi-scale filters for semantic segmentation.
\newblock In {\em ICCV}, 2019.

\bibitem{he2016deep}
Kaiming He, Xiangyu Zhang, Shaoqing Ren, and Jian Sun.
\newblock Deep residual learning for image recognition.
\newblock In {\em CVPR}, 2016.

\bibitem{He_2019_CVPR}
Tong He, Chunhua Shen, Zhi Tian, Dong Gong, Changming Sun, and Youliang Yan.
\newblock Knowledge adaptation for efficient semantic segmentation.
\newblock In {\em CVPR}, 2019.

\bibitem{he2017std2p}
Yang He, Wei-Chen Chiu, Margret Keuper, and Mario Fritz.
\newblock Std2p: Rgbd semantic segmentation using spatio-temporal data-driven
  pooling.
\newblock In {\em CVPR}, 2017.

\bibitem{hinton2015distilling}
Geoffrey Hinton, Oriol Vinyals, and Jeff Dean.
\newblock Distilling the knowledge in a neural network.
\newblock {\em arXiv preprint arXiv:1503.02531}, 2015.

\bibitem{huang2017densely}
Gao Huang, Zhuang Liu, Laurens Van Der~Maaten, and Kilian~Q Weinberger.
\newblock Densely connected convolutional networks.
\newblock In {\em CVPR}, 2017.

\bibitem{ioannou2017deep}
Yani Ioannou, Duncan Robertson, Roberto Cipolla, and Antonio Criminisi.
\newblock Deep roots: Improving cnn efficiency with hierarchical filter groups.
\newblock In {\em CVPR}, 2017.

\bibitem{jain2019accel}
Samvit Jain, Xin Wang, and Joseph~E Gonzalez.
\newblock Accel: A corrective fusion network for efficient semantic
  segmentation on video.
\newblock In {\em CVPR}, 2019.

\bibitem{jin2017video}
Xiaojie Jin, Xin Li, Huaxin Xiao, Xiaohui Shen, Zhe Lin, Jimei Yang, Yunpeng
  Chen, Jian Dong, Luoqi Liu, Zequn Jie, et~al.
\newblock Video scene parsing with predictive feature learning.
\newblock In {\em ICCV}, 2017.

\bibitem{fu2018dual}
Haijie Tian Yong Li Yongjun Bao Zhiwei Fang and Hanqing~Lu Jun~Fu, Jing~Liu.
\newblock Dual attention network for scene segmentation.
\newblock 2019.

\bibitem{kreso2017ladder}
Ivan Kreso, Sinisa Segvic, and Josip Krapac.
\newblock Ladder-style densenets for semantic segmentation of large natural
  images.
\newblock In {\em ICCV Workshop}, 2017.

\bibitem{krizhevsky2012imagenet}
Alex Krizhevsky, Ilya Sutskever, and Geoffrey~E Hinton.
\newblock Imagenet classification with deep convolutional neural networks.
\newblock In {\em NIPS}, 2012.

\bibitem{kundu2016feature}
Abhijit Kundu, Vibhav Vineet, and Vladlen Koltun.
\newblock Feature space optimization for semantic video segmentation.
\newblock In {\em CVPR}, 2016.

\bibitem{li2019dfanet}
Hanchao Li, Pengfei Xiong, Haoqiang Fan, and Jian Sun.
\newblock Dfanet: Deep feature aggregation for real-time semantic segmentation.
\newblock In {\em CVPR}, 2019.

\bibitem{li2019expectation}
Xia Li, Zhisheng Zhong, Jianlong Wu, Yibo Yang, Zhouchen Lin, and Hong Liu.
\newblock Expectation-maximization attention networks for semantic
  segmentation.
\newblock {\em ICCV}, 2019.

\bibitem{li2019attention}
Yanwei Li, Xinze Chen, Zheng Zhu, Lingxi Xie, Guan Huang, Dalong Du, and
  Xingang Wang.
\newblock Attention-guided unified network for panoptic segmentation.
\newblock In {\em CVPR}, 2019.

\bibitem{li2018low}
Yule Li, Jianping Shi, and Dahua Lin.
\newblock Low-latency video semantic segmentation.
\newblock In {\em CVPR}, 2018.

\bibitem{liu2019auto}
Chenxi Liu, Liang-Chieh Chen, Florian Schroff, Hartwig Adam, Wei Hua, Alan~L
  Yuille, and Li Fei-Fei.
\newblock Auto-deeplab: Hierarchical neural architecture search for semantic
  image segmentation.
\newblock In {\em CVPR}, 2019.

\bibitem{Liu_2019_CVPR}
Yifan Liu, Ke Chen, Chris Liu, Zengchang Qin, Zhenbo Luo, and Jingdong Wang.
\newblock Structured knowledge distillation for semantic segmentation.
\newblock In {\em CVPR}, 2019.

\bibitem{long2015fully}
Jonathan Long, Evan Shelhamer, and Trevor Darrell.
\newblock Fully convolutional networks for semantic segmentation.
\newblock In {\em CVPR}, 2015.

\bibitem{mahasseni2017budget}
Behrooz Mahasseni, Sinisa Todorovic, and Alan Fern.
\newblock Budget-aware deep semantic video segmentation.
\newblock In {\em CVPR}, 2017.

\bibitem{mazzini2018guided}
Davide Mazzini.
\newblock Guided upsampling network for real-time semantic segmentation.
\newblock {\em BMVC}, 2018.

\bibitem{Silberman12nyyd}
Pushmeet~Kohli Nathan~Silberman, Derek~Hoiem and Rob Fergus.
\newblock Indoor segmentation and support inference from rgbd images.
\newblock In {\em ECCV}, 2012.

\bibitem{Nekrasov_2020_WACV}
Vladimir Nekrasov, Hao Chen, Chunhua Shen, and Ian Reid.
\newblock Architecture search of dynamic cells for semantic video segmentation.
\newblock In {\em WACV}, 2020.

\bibitem{nilsson2018semantic}
David Nilsson and Cristian Sminchisescu.
\newblock Semantic video segmentation by gated recurrent flow propagation.
\newblock In {\em CVPR}, 2018.

\bibitem{oh2019video}
Seoung~Wug Oh, Joon-Young Lee, Ning Xu, and Seon~Joo Kim.
\newblock Video object segmentation using space-time memory networks.
\newblock {\em ICCV}, 2019.

\bibitem{orsic2019defense}
Marin Orsic, Ivan Kreso, Petra Bevandic, and Sinisa Segvic.
\newblock In defense of pre-trained imagenet architectures for real-time
  semantic segmentation of road-driving images.
\newblock In {\em CVPR}, 2019.

\bibitem{paszke2016enet}
Adam Paszke, Abhishek Chaurasia, Sangpil Kim, and Eugenio Culurciello.
\newblock Enet: A deep neural network architecture for real-time semantic
  segmentation.
\newblock {\em arXiv preprint arXiv:1606.02147}, 2016.

\bibitem{Paul_2020_WACV}
Matthieu Paul, Christoph Mayer, Luc~Van Gool, and Radu Timofte.
\newblock Efficient video semantic segmentation with labels propagation and
  refinement.
\newblock In {\em WACV}, 2020.

\bibitem{peng2017large}
Chao Peng, Xiangyu Zhang, Gang Yu, Guiming Luo, and Jian Sun.
\newblock Large kernel matters--improve semantic segmentation by global
  convolutional network.
\newblock In {\em CVPR}, 2017.

\bibitem{shelhamer2016clockwork}
Evan Shelhamer, Kate Rakelly, Judy Hoffman, and Trevor Darrell.
\newblock Clockwork convnets for video semantic segmentation.
\newblock In {\em ECCV}, 2016.

\bibitem{shuai2016dag}
Bing Shuai, Zhen Zuo, Bing Wang, and Gang Wang.
\newblock Dag-recurrent neural networks for scene labeling.
\newblock In {\em CVPR}, 2016.

\bibitem{simonyan2014very}
Karen Simonyan and Andrew Zisserman.
\newblock Very deep convolutional networks for large-scale image recognition.
\newblock {\em ICLR}, 2015.

\bibitem{szegedy2017inception}
Christian Szegedy, Sergey Ioffe, Vincent Vanhoucke, and Alexander~A Alemi.
\newblock Inception-v4, inception-resnet and the impact of residual connections
  on learning.
\newblock In {\em AAAI}, 2017.

\bibitem{Takikawa_2019_ICCV}
Towaki Takikawa, David Acuna, Varun Jampani, and Sanja Fidler.
\newblock Iccv.
\newblock 2019.

\bibitem{TangDPBS18}
Meng Tang, Abdelaziz Djelouah, Federico Perazzi, Yuri Boykov, and Christopher
  Schroers.
\newblock Normalized cut loss for weakly-supervised {CNN} segmentation.
\newblock In {\em CVPR}, 2018.

\bibitem{TangPDASB18}
Meng Tang, Federico Perazzi, Abdelaziz Djelouah, Ismail~Ben Ayed, Christopher
  Schroers, and Yuri Boykov.
\newblock On regularized losses for weakly-supervised {CNN} segmentation.
\newblock In {\em ECCV}, 2018.

\bibitem{tripathi2015semantic}
Subarna Tripathi, Serge Belongie, Youngbae Hwang, and Truong Nguyen.
\newblock Semantic video segmentation: Exploring inference efficiency.
\newblock In {\em ISOCC}. IEEE.

\bibitem{vaswani2017attention}
Ashish Vaswani, Noam Shazeer, Niki Parmar, Jakob Uszkoreit, Llion Jones,
  Aidan~N Gomez, {\L}ukasz Kaiser, and Illia Polosukhin.
\newblock Attention is all you need.
\newblock In {\em NIPS}, 2017.

\bibitem{veit2016residual}
Andreas Veit, Michael~J Wilber, and Serge Belongie.
\newblock Residual networks behave like ensembles of relatively shallow
  networks.
\newblock In {\em NIPS}, 2016.

\bibitem{wang2018non}
Xiaolong Wang, Ross Girshick, Abhinav Gupta, and Kaiming He.
\newblock Non-local neural networks.
\newblock In {\em CVPR}, 2018.

\bibitem{wu2019wider}
Zifeng Wu, Chunhua Shen, and Anton Van Den~Hengel.
\newblock Wider or deeper: Revisiting the resnet model for visual recognition.
\newblock {\em Pattern Recognition}, 2019.

\bibitem{xu2018dynamic}
Yu-Syuan Xu, Tsu-Jui Fu, Hsuan-Kung Yang, and Chun-Yi Lee.
\newblock Dynamic video segmentation network.
\newblock In {\em CVPR}, 2018.

\bibitem{yu2018bisenet}
Changqian Yu, Jingbo Wang, Chao Peng, Changxin Gao, Gang Yu, and Nong Sang.
\newblock Bisenet: Bilateral segmentation network for real-time semantic
  segmentation.
\newblock In {\em ECCV}, 2018.

\bibitem{Zagoruyko2016WRN}
Sergey Zagoruyko and Nikos Komodakis.
\newblock Wide residual networks.
\newblock In {\em BMVC}, 2016.

\bibitem{zhang2018context}
Hang Zhang, Kristin Dana, Jianping Shi, Zhongyue Zhang, Xiaogang Wang, Ambrish
  Tyagi, and Amit Agrawal.
\newblock Context encoding for semantic segmentation.
\newblock In {\em CVPR}, 2018.

\bibitem{zhao2018icnet}
Hengshuang Zhao, Xiaojuan Qi, Xiaoyong Shen, Jianping Shi, and Jiaya Jia.
\newblock Icnet for real-time semantic segmentation on high-resolution images.
\newblock In {\em ECCV}, 2018.

\bibitem{zhao2017pyramid}
Hengshuang Zhao, Jianping Shi, Xiaojuan Qi, Xiaogang Wang, and Jiaya Jia.
\newblock Pyramid scene parsing network.
\newblock In {\em CVPR}, 2017.

\bibitem{zhu2019empirical}
Xizhou Zhu, Dazhi Cheng, Zheng Zhang, Stephen Lin, and Jifeng Dai.
\newblock An empirical study of spatial attention mechanisms in deep networks.
\newblock {\em ICCV}, 2019.

\bibitem{zhu2017deep}
Xizhou Zhu, Yuwen Xiong, Jifeng Dai, Lu Yuan, and Yichen Wei.
\newblock Deep feature flow for video recognition.
\newblock In {\em CVPR}, 2017.

\bibitem{Zhu_2019_CVPR}
Yi Zhu, Karan Sapra, Fitsum~A. Reda, Kevin~J. Shih, Shawn Newsam, Andrew Tao,
  and Bryan Catanzaro.
\newblock Improving semantic segmentation via video propagation and label
  relaxation.
\newblock In {\em CVPR}, 2019.

\bibitem{Zhu_2019_ICCV}
Zhen Zhu, Mengde Xu, Song Bai, Tengteng Huang, and Xiang Bai.
\newblock Asymmetric non-local neural networks for semantic segmentation.
\newblock In {\em ICCV}, 2019.

\end{thebibliography}
